# Transcriptome-wide prediction of prostate cancer gene expression from histopathology images using co-expression based convolutional neural networks

Philippe Weitz[1], Yinxi Wang[1], Kimmo Kartasalo[1,2], Lars Egevad[3], Johan Lindberg[1,4], Henrik Grönberg[1], Martin Eklund[1], Mattias Rantalainen[1]

[1] Department of Medical Epidemiology and Biostatistics, Karolinska Institutet, Stockholm, Sweden
[2] Faculty of Medicine and Health Technology, Tampere University, Tampere, Finland
[3] Department of Oncology and Pathology, Karolinska Institutet, Stockholm, Sweden
[4] Science for Life Laboratory, Stockholm, Sweden

# Abstract

Molecular phenotyping by gene expression profiling is common in contemporary cancer research and in molecular diagnostics. However, molecular profiling remains costly and resource intense to implement, and is just starting to be introduced into clinical diagnostics. Molecular changes, including genetic alterations and gene expression changes, occuring in tumors cause morphological changes in tissue, which can be observed on the microscopic level. The relationship between morphological patterns and some of the molecular phenotypes can be exploited to predict molecular phenotypes directly from routine haematoxylin and eosin (H&E) stained whole slide images (WSIs) using deep convolutional neural networks (CNNs). In this study, we propose a new, computationally efficient approach for disease specific modelling of relationships between morphology and gene expression, and we conducted the first transcriptome-wide analysis in prostate cancer, using CNNs to predict bulk RNA-sequencing estimates from WSIs of H&E stained tissue. The work is based on the TCGA PRAD study and includes both WSIs and RNA-seq data for 370 patients. Out of 15586 protein coding and sufficiently frequently expressed transcripts, 6618 had predicted expression significantly associated with RNA-seq estimates (FDR-adjusted p-value $< 1*10^{-4}$) in a cross-validation. 5419 (81.9%) of these were subsequently validated in a held-out test set. We also demonstrate the ability to predict a prostate cancer specific cell cycle progression score directly from WSIs. These findings suggest that contemporary computer vision models offer an inexpensive and scalable solution for prediction of gene expression phenotypes directly from WSIs, providing opportunity for cost-effective large-scale research studies and molecular diagnostics.

# Introduction

Prostate cancer is one of the most common types of cancer and cause of cancer related deaths in men[1]. Molecular phenotyping is currently increasing in importance in both the research and clinical settings, as it enables detailed characterisation of individual tumours and provides information that enables cancer precision medicine[2]. Molecular phenotyping can reveal molecular etiology [3–5], predictive and prognostic markers[6,7], and enable molecular subtyping[8–10]. Gene expression profiling by RNA-sequencing offers a broad molecular phenotype of prostate cancer[11,12]. In recent years, several gene expression-based prostate cancer assays for clinical use have been introduced. The Prolaris cell-cycle progression score (CCP) provides an assessment of disease aggressiveness, a 10-year risk of metastasis after therapy, risk of recurrence after prostatectomy and disease-specific mortality under conservative management based on the mean mRNA expression of 31 genes in either biopsy or prostatectomy tissue[13–15]. Other mRNA-based diagnostic tests are the Oncotype Dx genomic prostate score[16–20], the Decipher Biopsy and Postoperative scores[21–23]. However, molecular phenotyping remains costly and time-consuming. There is therefore a demand for tools that can be used to cost-efficiently identify the molecular characteristics of large cohorts of patients retrospectively in research studies, as well as patients in the clinic. This has the potential to identify both novel biomarkers as well as help prioritising patients that may benefit from more comprehensive molecular phenotyping.

With the advent of digital pathology, where histopathology slides are digitized as part of the routine workflow, computer-based image analysis can now be applied to analyse morphological patterns in histopathology images. It has been demonstrated that computer vision models can be applied to predict molecular characteristics from tissue morphology, including mutations, molecular subtypes[24,25], and gene expression[26,27]. Compared with conventional bulk DNA- or RNA-sequencing, these models also capture spatially resolved intra-tumor heterogeneity[28,29]. While previous studies have demonstrated the feasibility to predict molecular phenotypes from H&E stained WSIs, the majority of these models are pan-cancer models[26,27] based on tumors originating from a range of organs. Although it can be assumed that some morphological patterns are shared among these tumors, it is unlikely that morphological patterns in general share their specific association with gene expression features across different cancers. Hence, cancer-specific models are almost certainly required to achieve optimal prediction performance.

To date no comprehensive analysis of the potential of computer vision models for whole-transcriptome analysis in prostate cancer has been reported. We therefore conducted a transcriptome-wide analysis of gene expression prediction modelling specifically for prostate cancer using data from the TCGA PRAD[11] study, applying a rigorous performance estimation strategy. We developed a novel computationally efficient modelling approach that exploits the co-expression patterns in gene expression data. This methodology can be deployed on relatively constrained computational infrastructure. Previous studies with this objective either relied on convolutional neural networks (CNNs) as feature extractors, with secondary models fitted to the CNN features, or on single transcript CNNs[28]. These approaches are either limited in capacity to learn domain-specific representations, or are computationally very costly. We therefore propose to jointly predict individual expressions in clusters of co-expressed (correlated) genes with multi-output models. This allows exploiting potential shared patterns and investigating the possibility of predicting transcripts and pathways that have previously been implicated in prostate cancer. To demonstrate a clinically relevant application, we show that this approach can be applied to predict the prognostic cell cycle progression score[13–15].

# Materials and methods

## Study materials

This study is based on image and expression data from the publicly available TCGA PRAD[11] data set which consists of 403 patients with 449 WSIs of formalin fixed paraffin embedded (FFPE) H&E stained sections of resected prostate tumors. These patients originate from 27 cancer centers and organisations, each of which contributed between 1 and 62 patients. From these 403 patients, 399 patients with adenomas and adenocarcinomas were included in this study, whereas 4 patients with ductal and lobular neoplasms were excluded. Of these patients, 389 with matching WSIs and gene expression data from tumor tissue available through TCGA were further selected. For patients with multiple WSIs available, we included one at random. A further 9 patients were excluded due to a prior systemic treatment or synchronous malignancies. The patient selection is shown in Supplementary Figure 1. We tiled and preprocessed the WSIs of these 380 cases as described in the Supplementary Materials and Methods, and identified cancer regions using a cancer detection model that we developed with transrectal core needle biopsy data from the STHLM3 prostate cancer diagnostic study[30,31] (see Supplementary Materials and Methods). Subsequently, 10 patients whose largest contiguous tumor area was below 1 mm² were excluded. The remaining 370 patients were included in this study. For each WSI, we only included tiles that we predicted to be malignant. We then randomly selected 92 (24.86%) of these 370 patients as a held-out test set. To this end, we computed 500 random splits stratified by ISUP grade and selected the split with the best matching age distributions as determined by a Kolmogorov-Smirnov test. The remaining 278 patients, which we will refer to as the development set, were further split into 10 cross-validation (CV) folds. The demographic and clinical characteristics of both the cross-validation set and the held-out test set are provided in Table 1.

## Gene selection

The TCGA PRAD RNA expression data includes 60843 transcripts. The biomaRt[32] *hsapiens_gene_ensembl* currently lists 22802 transcripts as protein coding. Of these, expression levels were available for 19601 transcripts in the TCGA PRAD data set, which were selected for further analyses. We only included genes for which there are at least three counts in at least 10% of patients, since less frequently expressed genes may not be possible to model with the number of samples in this study. This further excludes 4015 transcripts, resulting in a set of 15586 included transcripts. In subsequent analyses, normalized expression values were used (log2 of the upper quartile normalised fragments per kilobase of transcript per million mapped reads (FPKM-UQ) as preprocessed with HTSeq[33]).

## Identification of sets of co-expressed transcripts

In order to reduce the computational complexity of predicting expression levels of 15586 transcripts, we propose a novel approach based on clustering the transcripts based on their co-expression. Clustered transcripts were subsequently jointly predicted with multi-output CNN models, with the expression values of the transcripts in each cluster as the response variables, such that a cluster consisting of *n* transcripts is predicted by a CNN with *n* outputs, one for each transcript. For the clustering, we defined the distance between two genes $x, y$ as $d = 1 - |\rho(x, y)|$, $d \in [0, 1]$, where $\rho(x, y)$ denotes the Spearman correlation between $x, y$. Clustering was performed in R with *hclust* and *ward.D* as the agglomeration method. As a compromise between computational efficiency, which requires few clusters to reduce the number of required CNNs, and high average intra-cluster correlation, we chose to cluster the 15586 selected transcripts into 50 clusters. Figure 1 depicts the number of transcripts initially included in each cluster, the number of transcripts in each cluster brought forward for validation in the test set, and the average absolute Spearman correlation for all gene pairs within the clusters. Supplementary Figure 4 shows the average absolute within- and across cluster Spearman correlation for all gene pairs within a cluster or between two clusters.

## Model optimisation and performance evaluation

We compared the joint cluster prediction with three alternative modelling approaches. In the first, we optimised a CNN to jointly predict the expression of all 15586 genes in one single model. In the second approach, we extracted a feature vector for each tile using an ImageNet[34] pretrained ResNet18[35] model and fitted boosting models with *LightGBM*[36] (lgbm) to predict gene expression with one boosting model per gene. To reduce the computational cost during model selection, these models were compared in a randomly selected subset of 10 clusters that contains 2636 transcripts. As a final baseline, we randomly reassigned all genes of this subset into 10 random clusters of matched sizes to investigate whether representations learned with the combination of gradients of co-expressed genes yield improved performance compared to a random combination.

For each transcript, we centered and scaled all expression values by the mean and variance of the respective training data before training the corresponding model. We then assigned this slide level expression as response for all the tiles of the respective slide. The mean prediction across tiles was used to assign a slide level predicted expression value. In order to preserve the independence of the validation folds and to reduce computational cost, hyperparameters were tuned in 5 different fold allocations for this subset of 10 clusters for the CNN models, whereas hyperparameters of the single-gene lgbm models were optimised on a random subset of 200 out of these 2636 transcripts. This validation procedure is comparable to a nested 10-fold cross-validation, as shown in Supplementary Figure 2. For 5 of the 10 splits, we excluded the respective outer validation fold, used 2 folds as inner validation folds and 7 folds for model fitting. Predictions for each of the validation folds were concatenated to obtain an independent prediction for each patient in the development set.

All models were optimised with a mean-squared-error loss function. To compensate for variability in the absolute cancer tissue area across slides and associated number of tiles *m*, we assigned a sampling weight of $w = log(1 + m)$ to each tile during training. Based on this, we randomly sampled 24,000 tiles with replacement per training epoch. Images were augmented with random rotations, random mirroring and random cropping to an image size of 440 x 440 pixels. We used four Nvidia RTX 2080Ti GPUs with mixed precision, using parallel single-gpu training with a batch size of 144 samples per GPU. As a model architecture, we selected ResNet18 initialized with ImageNet weights. Both the considered and selected hyperparameters for the CNNs and the boosting model are shown in Supplementary Tables 2 and 3.

After selecting the best performing out of the four investigated modelling approaches based on their performance on the outer validation folds, we performed the inner cross-validation with the 40 remaining clusters with the best performing model to determine an optimal resolution out of 40X, 20X and 10X for each of the remaining transcripts. We then fitted one CNN for each cluster and resolution level with 9 training folds that include the 2 inner validation folds and the prediction performance of each of the 15586 transcripts was evaluated on the respective outer validation folds. While each cluster was predicted entirely at every considered resolution level, we only used the prediction at the resolution level that was previously determined as optimal for the respective transcript.

Spearman rank correlations between the slide level predictions and the RNA-seq expression level values were used as the primary performance metric. Genes with FDR[37] adjusted p-value <0.0001 were brought forward for validation in the test data. To obtain predictions in the test data, we predicted all test set tiles with all 500 models from the 10 folds and 50 clusters for all 3 resolution levels, and averaged over the 10 predictions (one from each of the 10 cross-validation models) per tile at the resolution level that was selected for each gene.

## Gene set enrichment analysis

Gene set enrichment analysis[38] (GSEA) was applied to investigate whether any specific biological functions were implicated with transcripts that were associated with morphology. The Reactome[39] pathway knowledge database was used in the analysis together with genes (15586) ranked by their p-values. GSEA was performed on p-values from the cross-validation data rather than the test data, since ranked enrichment analysis can identify significantly enriched gene sets even if a proportion of the included genes did not meet any significance thresholds.

## Cell-cycle progression score

In order to investigate potential clinical applications of this modeling approach, we computed the cell cycle progression score[13–15], both from the TCGA RNA-seq expression data and from model predictions. The CCP is a commercial prognostic test that is intended to support clinical decision making and is computed by taking the mean of 31 highly correlated gene expression levels. We evaluated the prediction performance by computing an RNA-seq based CCP and assessed the Spearman correlation between this score and a CNN-based score that was computed as the mean of all CCP genes that met the validation criterion for the test set (FDR-adjusted p-value < 0.0001 in the CV data). In order to evaluate whether the prognostic performance of the CNN-predicted CCP is comparable to the CCP based on the RNA-seq data, we performed univariate hazard analysis with Cox proportional hazard models with time to biochemical recurrence as the outcome.

# Results

We developed and applied a new approach for transcriptome-wide prediction of prostate cancer gene expression using deep CNN models. Prediction performance was validated in a held-out test set.

## Evaluation of modelling strategies

We first evaluated four CNN-based modelling approaches for the prediction of gene expression in a subset of 2636 transcripts from 10 randomly drawn clusters (see Methods). The cluster-based approach, which exploits shared representations for co-expressed genes, achieved the highest average Spearman correlation (0.243) as well as the highest number (1191 out of 2636, 45.18%) of significant correlations (FDR adjusted p-values <0.0001). Predicting genes in randomly assigned clusters resulted in 1030 (39.07%) significant correlations. Fitting lgbm boosting models to ImageNet ResNet18 features with one boosting model per gene or predicting all selected 15586 genes jointly with a single CNN resulted in 693 (26.29%) and 0 (0%) significant correlations out of 2636 genes respectively. The distribution of Spearman correlations for each modelling approach is visualised in Figure 2a, with corresponding adjusted p-values shown Figure 2b. Average compute time per gene was also assessed (Figure 2c), revealing substantially shorter compute time for the cluster-based approach compared to fitting secondary models to extracted features.

## Transcriptome-wide prediction of prostate cancer expression values

Based on the model comparison in the previous section, the cluster-based method was selected for the transcriptome-wide analysis across all 15,586 transcripts. First, the prediction performance across all transcripts was assessed in (nested) cross-validation (Figure 2a). Out of the 15586 predicted gene expression levels, 6618 (42.5%) were associated with the corresponding RNA-sequencing based estimates (Spearman correlation, FDR adjusted p-value < $1*10^{-4}$, adjustment with the method described by Benjamini and Hochberg (BH)[37]). The 6618 significant transcripts were brought forward for validation in the held-out test (92 patients). Out of the 6618 transcripts, 5419 (81.9%) had a BH-adjusted p-value <0.01 in the test set. The distributions of Spearman correlations are depicted in Figure 2c and d for the entire CV data and test set respectively. The gene with the highest Spearman correlation between RNA-seq and CNN prediction in the test set was BRICD5, with a correlation of 0.749. Figure 3 a shows scatter plots for the gene BRICD5 together with example tiles with low and high predicted expressions (Figure 4 b-c). BRICD5 belongs to the BRICHOS family, which is assumed to act as a chaperone in protein folding[40].

## Genes associated with molecular mechanisms of prostate cancer

Among the 5419 significantly predicted transcripts, several of the corresponding genes have previously been reported to be associated with molecular mechanisms of prostate cancer. Out of the 20 genes included in an expression-based androgen receptor activity score[11], three were significantly predicted from WSIs: GNMT, MPHOSPH9 and ZBTB10 with respective correlations of 0.51, 0.324 and 0.279. The relationship between predicted and RNA-seq expression estimates for GNMT is shown in Figure 3d, with examples of low and high expression in Figure 3e and f. Further significantly predicted genes in the androgen signalling pathway were NCOR1 (0.468), the gene encoding the androgen receptor (AR, 0.322) and NCOA2 (0.31), which has previously been found to be overexpressed in 8% of primary tumors and 37% of metastases[5]. FOXA1 and SPOP expression predictions were not significantly associated with their expression (Spearman correlations of 0.013 and 0.22 in CV). However, a human paralog of SPOP, SPOPL, which can act as a negative regulator of SPOP[41] was correlated with 0.526.

Expression of the DNA repair genes CDK12 (examples in Figure 3g-i), which is frequently mutated in metastatic prostate cancer[42], and ATM show Spearman correlations of 0.577 and 0.56 between predicted and RNA-seq expression. The human breast cancer genes BRCA1 and BRCA2 also belong to this functional group and their expression could be predicted with almost identical correlations of 0.49. The DNA mismatch repair genes MSH2 and MSH6 (0.383 and 0.305) have been found to be frequently mutated in hypermutated microsatellite unstable advanced prostate cancers[43]. Further significantly predicted DNA repair genes are FANCD2 and RAD51C with correlations of 0.395 and 0.335.

While PTEN did not meet the inclusion criterion due to low expression, multiple established tumor suppressor genes had a significant association between RNA-seq estimates of gene expression and prediction. ZFHX3, which could be predicted with a correlation of 0.6, is a tumor suppressor gene that downregulates proliferation via MYC in prostate cancer[44]. Other significantly associated tumor suppressor genes include APC, Rb1, KMT2D and KMT2C, with Spearman correlations of 0.6, 0512, 0.512 and 0.484.

The PI3K pathway is upregulated in 30-50% of prostate cancers and has been identified as a therapeutic target[45]. PIK3CA, and PIK3R1 were predicted with Spearman correlations of 0.458 and 0.407. The GTPase HRAS is upstream of the PI3K pathway and has a Spearman correlation of 0.568. MED12 is a subunit of the Mediator kinase complex and is essential in the transcription of protein coding genes. It is frequently over-expressed in castration resistant distant metastatic and locally recurrent prostate cancers as compared to androgen-sensitive prostate cancers or benign prostatic tissue[46] and could be predicted with a Spearman correlation of 0.454.

## Gene set enrichment analysis

Gene set enrichment analysis revealed 12 significantly enriched pathways that belong to the functional groups of the cell cycle, RNA metabolism, the immune system, the metabolism of proteins, signal transduction, hemostasis, chromatin organisation, the circadian clock and metabolism. Brief description of the identified pathways, their adjusted p-values as well as the distribution of Spearman correlations between CNN predictions and sequenced expression levels are depicted in Figure 4. The most significantly enriched pathway, R-HSA-113510 with an adjusted p-value of 0.005, regulates DNA replication through the Rb1 E2F pathway. This pathway has previously been found to be frequently mutated in prostate cancer[42]. Besides the tumor suppressor gene Rb1, this pathway also contains the CCP gene RRM2, which encodes a reductase that catalyzes the formation of deoxyribonucleotides from ribonucleotides. Both the second and third most strongly associated pathways, R-HSA-6782315 and R-HSA-72200, serve the metabolism of RNA. R-HSA-6782315, with an adjusted p-value of 0.07, is involved in tRNA modification in the nucleus and cytosol and has previously been implicated in human diseases, including cancer[47].

## Cell cycle progression score

Of the 31 genes that comprise the CCP, 29 were validated in the test set, which excludes CDC2 and CENPM. We therefore computed a CNN based CCP score as the average of the 29 remaining CCP genes and compared it with an RNA based CCP score that is based on all 31 transcripts. The Spearman correlations between the 29 CNN predictions and their RNA expression is depicted in Figure 5a and provided in Supplementary Table 1. The CNN CCP score has a Spearman correlation of 0.527 (bootstrapped 95% CI 0.357, 0.665) with its RNA-seq counterpart (Figure 3j, examples of low and high expression in Figure 3k-l). The corresponding area under the receiver operating characteristic curve (AUC) for classifying whether the CCP is expressed above or below its median in the test set is 0.733. Figure 5b reveals a comparable relationship between ISUP grade and ranked CCP score both for the CNN prediction and RNA-seq. Biochemical recurrence (BCR) is the only outcome with a sufficient number of events for time-to-event analysis in the TCGA PRAD study, with 50 (18%) and 20 (21.7%) patients with BCR events in the CV and the test set, respectively. The HR of the RNA-seq -based CCP was 1.68 (1.256, 2.246) in the CV and 1.351 (0.956, 1.909) in the test data. For the CNN predicted CCP, the respective HR values were 2.579 (1.412, 4.713) and 2.943 (1.055, 8.212)(Figure 5c). Figure 5d depicts CNN CCP predictions overlayed over representative example WSIs for cases of all ISUP grades.

# Discussion

In this study, we performed the first transcriptome-wide gene expression prediction specifically for prostate cancer and identified a set of 5419 genes whose expression is associated with morphological changes that are detectable by current computer vision models in the TCGA PRAD dataset. We furthermore evaluated this approach to predict a prognostic gene expression-based proliferation score. To this end, we optimised CNN models to predict 15886 frequently expressed protein coding genes and assessed four different computationally efficient modelling approaches.

As compared to fitting one CNN per gene, the co-expression -based modelling approach proposed here reduces the number of models that need to be fitted from 15586 to 50, which roughly translates to a 300-fold reduction in computational cost. This increases computational efficiency substantially and reduces hardware requirements and costs. Using correlated instead of randomly assigned clusters for joint prediction proved to be a computationally inexpensive way to increase model performance. We speculate that this may be because co-expression of genes is more likely to be associated with similar morphological features and therefore, representations learned in correlated clusters generalize across genes in each cluster.

Previous studies reported prediction of mRNA expression from WSIs of H&E stained tissue with pan-cancer models, including in the TCGA PRAD cohort[26,27]. In order to compare these to the results obtained in this study, we replicated the respective analysis as closely as possible. The study presented by Schmauch et al. is difficult to compare to this study since it only relies on cross-validation to assess prediction performance and reports Pearson correlation as the performance metric. Furthermore, the presented results include transcripts that are not known to encode proteins. Generally, the numbers of significantly predicted transcripts are in a similar order of magnitude. A direct comparison to the results by Fu et al. reveals a similar number of significantly predicted genes in the TCGA PRAD cohort. The difference in the number of significantly predicted transcripts could be explained through a different patient sample in the respective test sets or slightly improved model performance with the proposed method. Interestingly, the intersection of the sets of significantly correlated gene expressions is only 4203 genes. A possible explanation for this difference in prediction performance for the approximately 3500 remaining transcripts may be that the higher number of cases in the training set when training pan-cancer models across all cancers in TCGA aids the prediction of some transcripts whose expression is associated with similar morphological patterns across cancers but impairs prediction performance for transcripts that may require cancer-specific models.

This study has a few limitations. Although our results are based on data from a multi-center study and while we applied a stringent validation approach with both a fully independent internal test set and a nested cross-validation for model selection, we have not been able to perform validation in a fully independent cohort, since there are currently no additional studies available with both RNA-sequencing data and WSIs. The size of the current study is expected to be a limitation with respect to optimizing the models. We expect model performance to improve with more data both for already significantly predicted transcripts as well as with respect to the number of transcripts that can be predicted accurately. However, there are unknown upper limits to the correlations in this study since the tissue material used for bulk sequencing is not necessarily identical to the tissue sectioned and stained for the WSIs. This limits the correlations both due to noise in labels during training as well as when comparing predicted gene expression to bulk sequencing estimates. We based our models and predictions on regions of high tumor purity by identifying cancer regions with a cancer detection model. However, since the detection model was developed on biopsy data, it required additional calibration in the prostatectomy WSIs and we expect that cancer detection could potentially be improved further.

In the set of genes that were significantly predicted in this study, there were many genes that are implicated in prostate cancer. Particularly the expression of genes of the cell cycle and of genes involved in proliferation such as the genes of the CCP score were predicted significantly. Transcripts of known tumor suppressor and DNA repair genes CDK12, ATM,

BRCA1, BRCA2, Rb1, KMT2D and ZFHX3 were also predicted with high correlations. However, a surprisingly low number of genes from the androgen signalling pathway had a significant correlation between prediction and gene expression, despite the central role of androgen in prostatic carcinogenesis, with the exception of GNMT and a few other genes. Based on this, we can speculate that gene expression activity in the androgen signalling pathway has limited impact on tissue morphology. We identified 12 pathways that are enriched for genes that could be predicted from WSIs, including those related to cell cycle, metabolism of RNA and proteins, the immune system and signal transduction based on ranked gene set enrichment analysis. Some of these pathways had previously been implicated in prostate cancer. Further investigation into the relationship between the differential expression of the significantly correlated genes and their associated morphology may yield novel biological insight or candidates for diagnostic, prognostic or predictive biomarkers. Potential clinical use of computer vision based gene expression prediction was investigated through an analysis of the prognostic cell cycle progression score. Rank-based analysis revealed that the predicted CCP score has a similar relationship to the ISUP grade as the sequencing based score. Univariate time-to-event analysis with BCR as outcome revealed that both the RNA-seq -based and the CNN predicted CCP were prognostic in the CV analysis, whereas only the CNN predicted CCP was prognostic in the test set. This analysis was, however, based on a relatively low number of events and patients. Prediction of molecular phenotypes and cell cycle score from histopathology images may prove clinically useful in low-resource environments in which molecular diagnostics are unavailable, or to analyse large cohorts of patients for which sequencing is too costly, including large scale studies of archived slides that may not be suitable for RNA sequencing.

In conclusion, our findings indicate that the expression of a large number of genes is significantly associated with morphological patterns. While considering the limitation that only approximate prediction of gene expression levels is possible from histopathology images , this study provides further evidence of a strong association between routine clinical H&E stained histopathology slides and average tumor gene expression. We conclude that contemporary computer vision models offer an inexpensive and scalable solution for prediction of gene expression phenotypes directly from WSIs, providing opportunity for cost-effective large-scale research studies and molecular diagnostics.

# Figures and tables

## Table 1

| | **Development (N=278)** | **Test (N=92)** | **Overall (N=370)** |
|---|---|---|---|
| **Age** | | | |
| Mean (SD) | 61.2 (6.75) | 61.3 (7.07) | 61.2 (6.82) |
| Median [Min, Max] | 61.4 [43.5, 77.9] | 62.0 [42.0, 78.6] | 61.4 [42.0, 78.6] |
| Not reported | 8 (2.9%) | 1 (1.1%) | 9 (2.4%) |
| **PSA** | | | |
| Mean (SD) | 1.02 (3.80) | 0.801 (2.80) | 0.967 (3.58) |
| Median [Min, Max] | 0.100 [0, 37.4] | 0.100 [0, 15.6] | 0.100 [0, 37.4] |
| Not reported | 32 (11.5%) | 13 (14.1%) | 45 (12.2%) |
| **ISUP** | | | |
| 1 | 28 (10.1%) | 10 (10.9%) | 38 (10.3%) |
| 2 | 85 (30.6%) | 27 (29.3%) | 112 (30.3%) |
| 3 | 63 (22.7%) | 21 (22.8%) | 84 (22.7%) |
| 4 | 29 (10.4%) | 9 (9.8%) | 38 (10.3%) |
| 5 | 73 (26.3%) | 25 (27.2%) | 98 (26.5%) |
| **Stage** | | | |
| I | 86 (30.9%) | 25 (27.2%) | 111 (30.0%) |
| II | 42 (15.1%) | 14 (15.2%) | 56 (15.1%) |
| III | 11 (4.0%) | 5 (5.4%) | 16 (4.3%) |
| IV | 29 (10.4%) | 13 (14.1%) | 42 (11.4%) |
| Not Reported | 110 (39.6%) | 35 (38.0%) | 145 (39.2%) |
| **Clinical T** | | | |
| T1, T1a, T1b, T1c | 95 (34.2%) | 35 (38.0%) | 130 (35.1%) |
| T2, T2a, T2b, T2c | 103 (37.1%) | 32 (34.8%) | 135 (36.5%) |
| T3, T3a, T3b | 17 (6.1%) | 8 (8.7%) | 25 (6.8%) |
| T4 | 1 (0.4%) | 1 (1.1%) | 2 (0.5%) |
| Not Reported | 62 (22.3%) | 16 (17.4%) | 78 (21.1%) |
| **Pathologic T** | | | |
| T2, T2a, T2b, T2c | 113 (40.6%) | 29 (31.5%) | 142 (38.4%) |
| T3, T3a, T3b | 157 (56.5%) | 59 (64.1%) | 216 (58.4%) |
| T4 | 7 (2.5%) | 1 (1.1%) | 8 (2.2%) |
| Not Reported | 1 (0.4%) | 3 (3.3%) | 4 (1.1%) |
| **Clinical M** | | | |
| M0 | 251 (90.3%) | 84 (91.3%) | 335 (90.5%) |
| M1a | 1 (0.4%) | 0 (0%) | 1 (0.3%) |
| M1c | 0 (0%) | 1 (1.1%) | 1 (0.3%) |
| Not Reported | 26 (9.4%) | 7 (7.6%) | 33 (8.9%) |
| **Pathologic N** | | | |
| N0 | 199 (71.6%) | 55 (59.8%) | 254 (68.6%) |
| N1 | 37 (13.3%) | 18 (19.6%) | 55 (14.9%) |
| Not Reported | 42 (15.1%) | 19 (20.7%) | 61 (16.5%) |
| **Vital status** | | | |
| Alive | 272 (97.8%) | 88 (95.7%) | 360 (97.3%) |
| Dead | 4 (1.4%) | 4 (4.3%) | 8 (2.2%) |
| Not Reported | 2 (0.7%) | 0 (0%) | 2 (0.5%) |
| **Biochemical recurrence** | | | |
| BCR | 228 (82.0%) | 72 (78.3%) | 300 (81.1%) |
| no BCR | 50 (18.0%) | 20 (21.7%) | 70 (18.9%) |

Figure 1

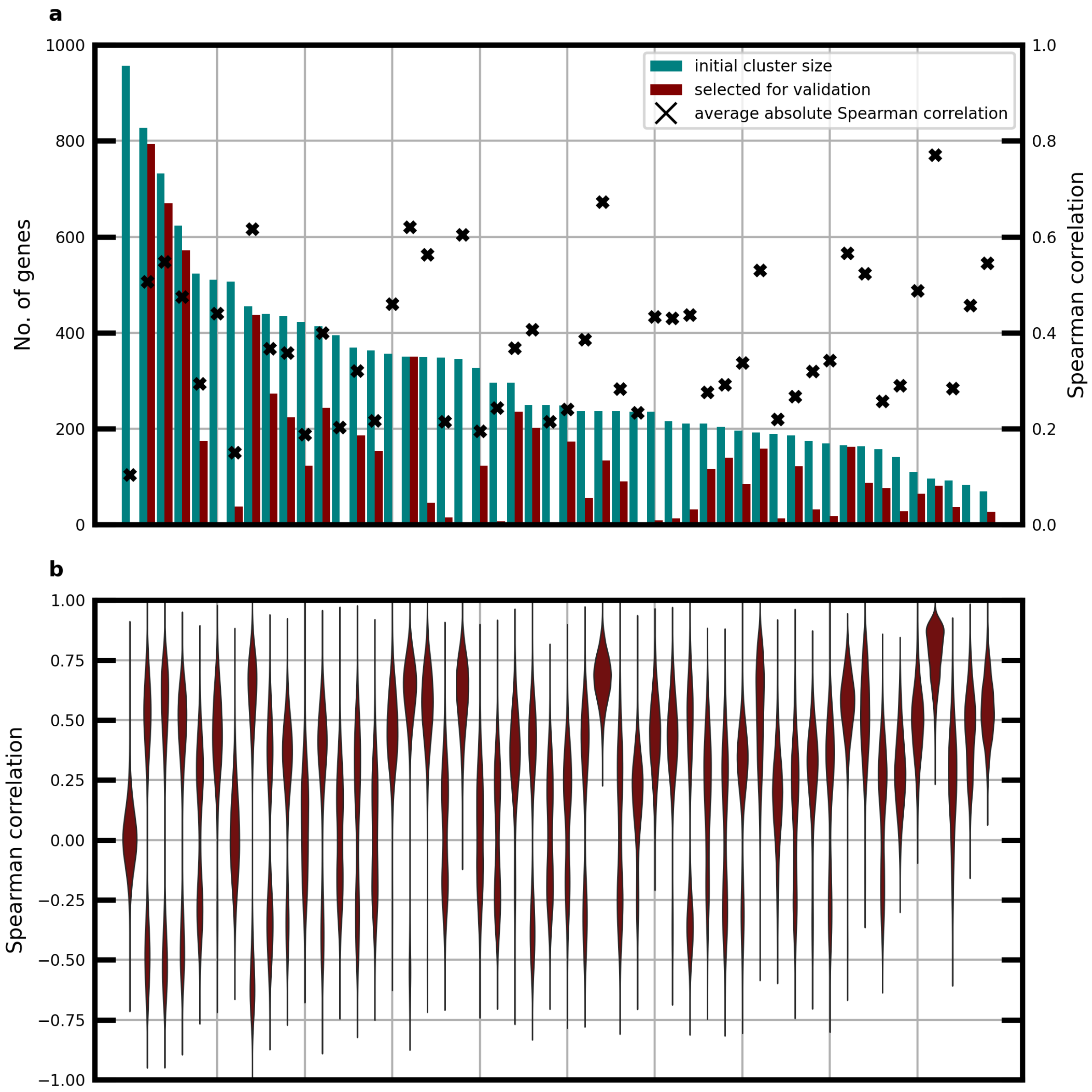

**Figure 1. a)** Number of genes per cluster after initial clustering using hierarchical clustering with Ward's method are depicted in green. The number of genes in each cluster that were significant in the cross-validation are depicted in red (FDR adjusted p-value of the Spearman correlation between gene expression and CNN prediction $<0.0001$). This corresponds to the number of genes that were further evaluated on the test data. Black crosses indicate the average absolute pairwise Spearman correlation between all genes in the respective cluster. **b)** The corresponding distribution of Spearman correlations between all pairs of genes within each cluster.

# Figure 2

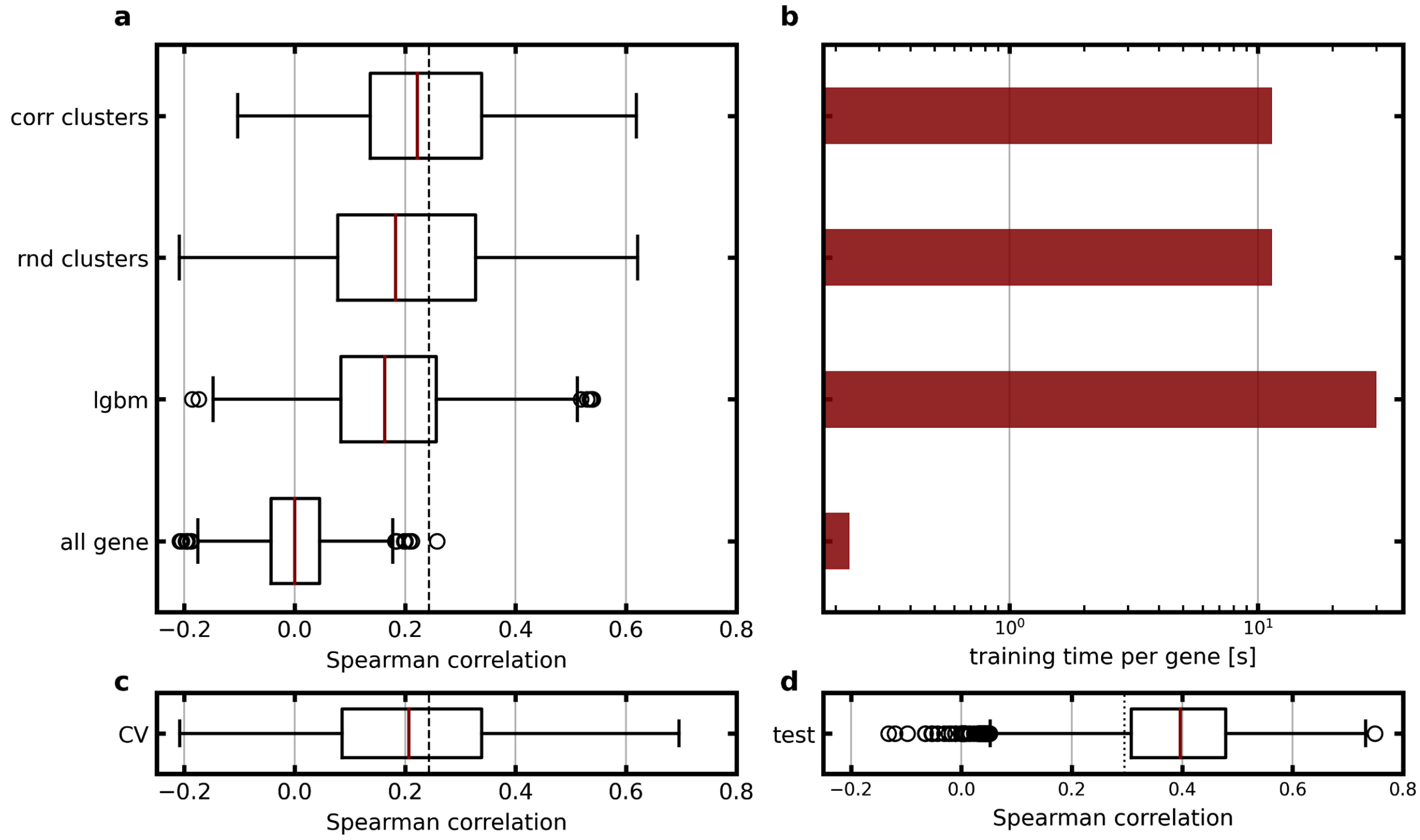

**Figure 2.** Boxplots of distributions of Spearman correlations for different modelling approaches and validation sets, as well as a comparison of computational efficiency. Boxes indicate the interquartile range, with the median marked by a red line. Whiskers extend an additional interquartile range from the quartiles, with points outside this interval marked with circles. Vertical dashed lines indicate the significance threshold for adjusted p-values of 0.0001 in the validation set, vertical dotted lines indicate the corresponding threshold in the test set of 0.01. **a)** Distributions of the Spearman correlations between CNN predictions and sequencing expression levels for 2636 genes from 10 randomly selected clusters. *corr clusters* refers to correlation based clustering, *rnd clusters* to random cluster assignments, *lgbm* to prediction with boosting models based on ResNet18 features and *all gene* to a cnn that predicts all 15586 selected genes at once (distribution shown only includes compared 2636 genes). For *corr clusters*, *rnd clusters*, *lgbm* and *all gene*, 1191, 1030, 693 and 0 genes had an FDR adjusted p-value lower than 0.0001 respectively. **b)** Average training time per gene for the different modelling approaches. **c)** Boxplot of Spearman correlations between gene expression and the respective CNN prediction for all 50 clusters comprising 15586 genes in the validation data, using the *corr clusters* method. 6618 genes had an adjusted p-value lower than 0.0001. **d)** Boxplot of Spearman correlations of the 6618 selected genes in the held-out test set, with 5419 adjusted p-values below 0.01.

# Figure 3

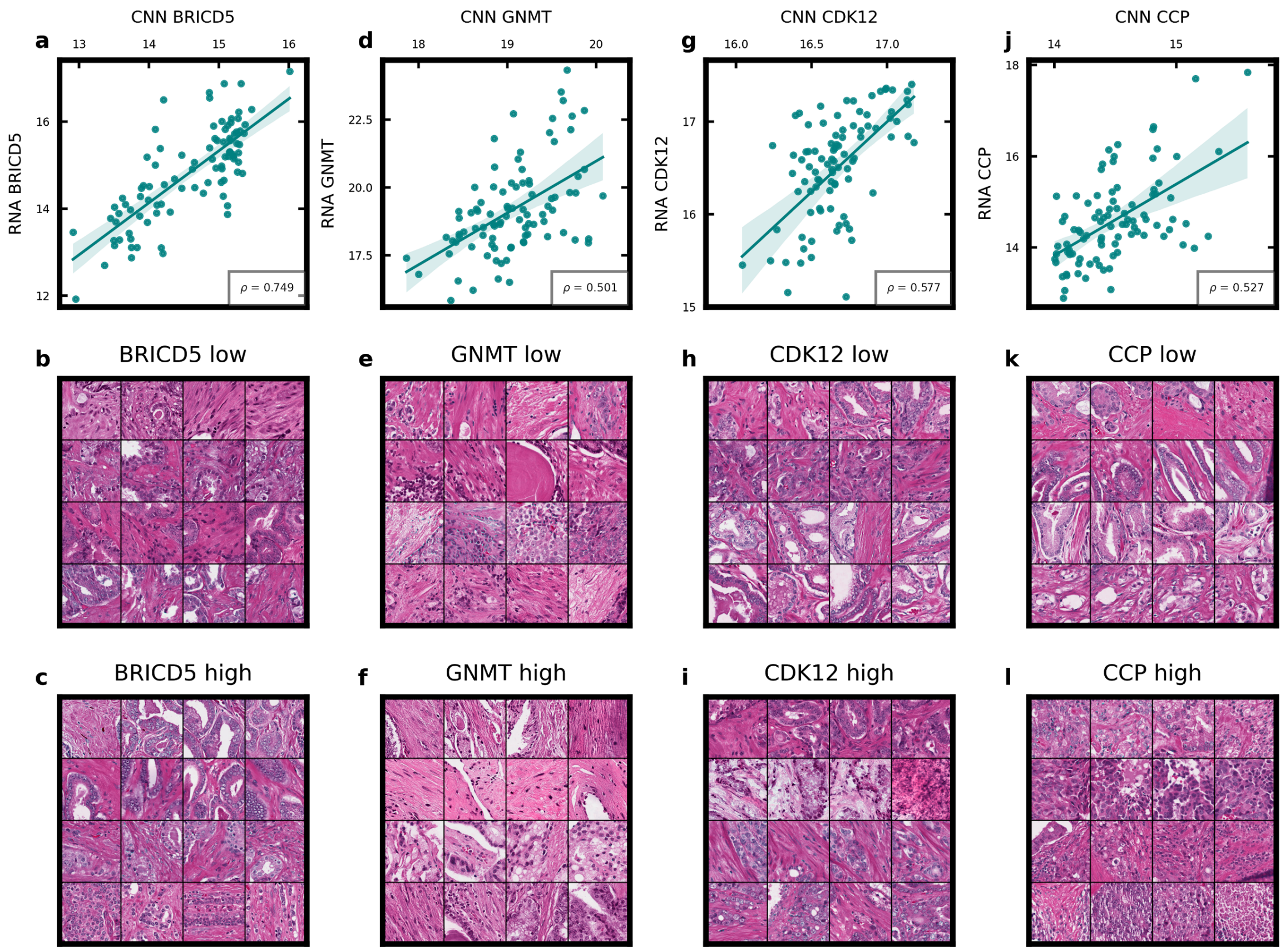

**Figure 3.** Comparison between predicted and RNA-seq expression, together with examples of tiles with low and high predicted expression for selected genes, with four patients with four tiles each per gene and expression level. For each subplot, the four tiles in the same row are sampled from the same WSI. The edge length of each of the 16 tiles is 110.88 µm. **a)** Scatter plot between CNN prediction and RNA-seq estimates of expression for the best predicted gene BRICD5 with a Spearman correlation of 0.749. **b)** Examples of tiles with low predicted BRICD5 expression. **c)** Example tiles with high predicted expression. **d)-f)** Corresponding plots for GNMT with a Spearman correlation of 0.501. GNMT is part of the androgen signalling pathway. **g)-i)** The respective relationship and examples for the DNA repair gene CDK12, with a Spearman correlation of 0.577. The corresponding plots for the CCP score are displayed in **j)-l)**, with higher expression being associated with higher proliferation, ISUP grade and poorer prognosis.

# Figure 4

| Group | Description |
|---|---|
| Cell Cycle | E2F mediated regulation of DNA replication |
| Metabolism of RNA | tRNA modification in the nucleus and cytosol |
| Metabolism of RNA | mRNA Editing: C to U Conversion |
| Immune System | FCGR activation |
| Metabolism of Proteins | Glycoprotein hormones |
| Signal Transduction | Antagonism of Activin by Follistatin |
| Hemostasis | Platelet Adhesion to exposed collagen |
| Immune System | Interleukin-17 signaling |
| Chromatin Organization | PKMTs methylate histone lysines |
| Circadian Clock | RORA activates gene expression |
| Metabolism | PPARA activates gene expression |
| Signal Transduction | Formyl peptide receptors bind formyl peptides |

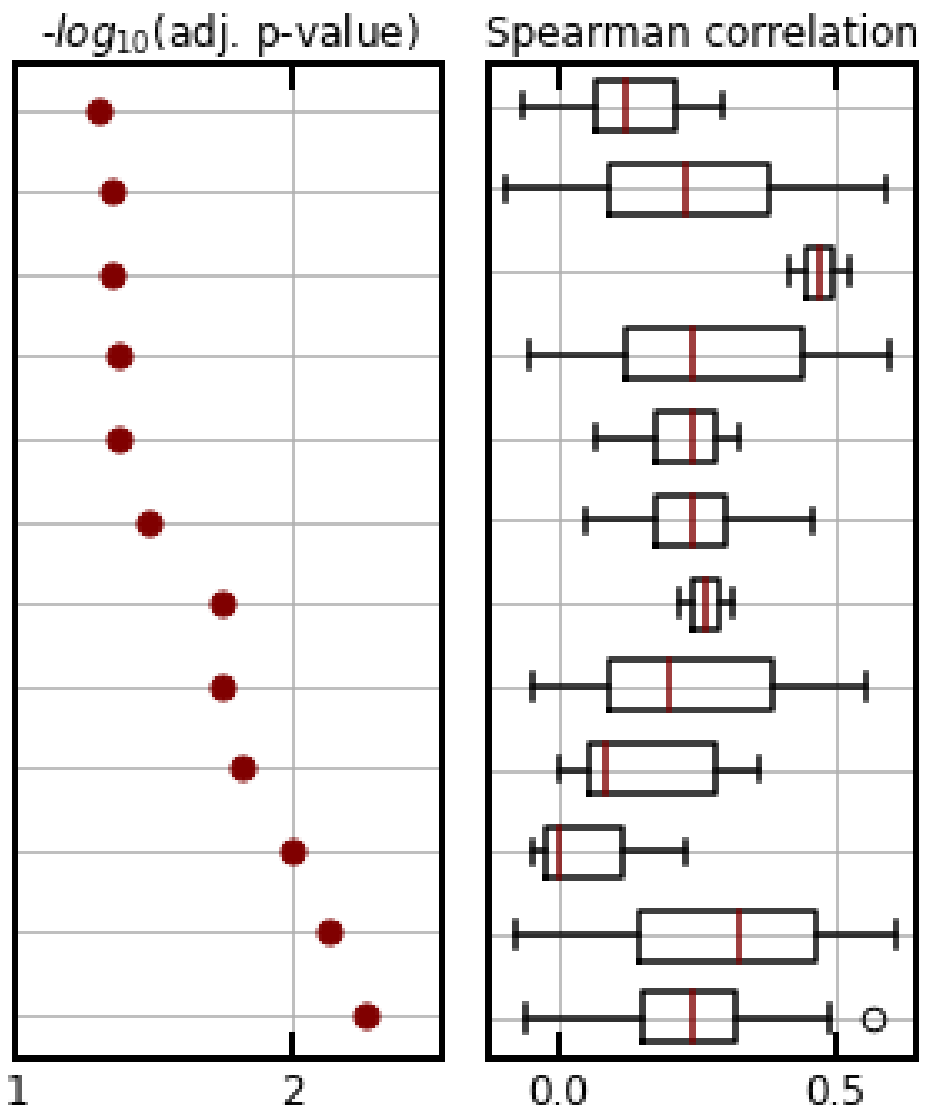

**Figure 4.** The12 significantly enriched pathways identified by gene set enrichment analysis with corresponding adjusted p-values from the cross-validation per pathway. The boxplots indicate the distribution of Spearman correlations between sequenced and predicted gene expression for all genes in the respective pathway. Boxplots are created with the same procedure as in Figure 2.

# Figure 5

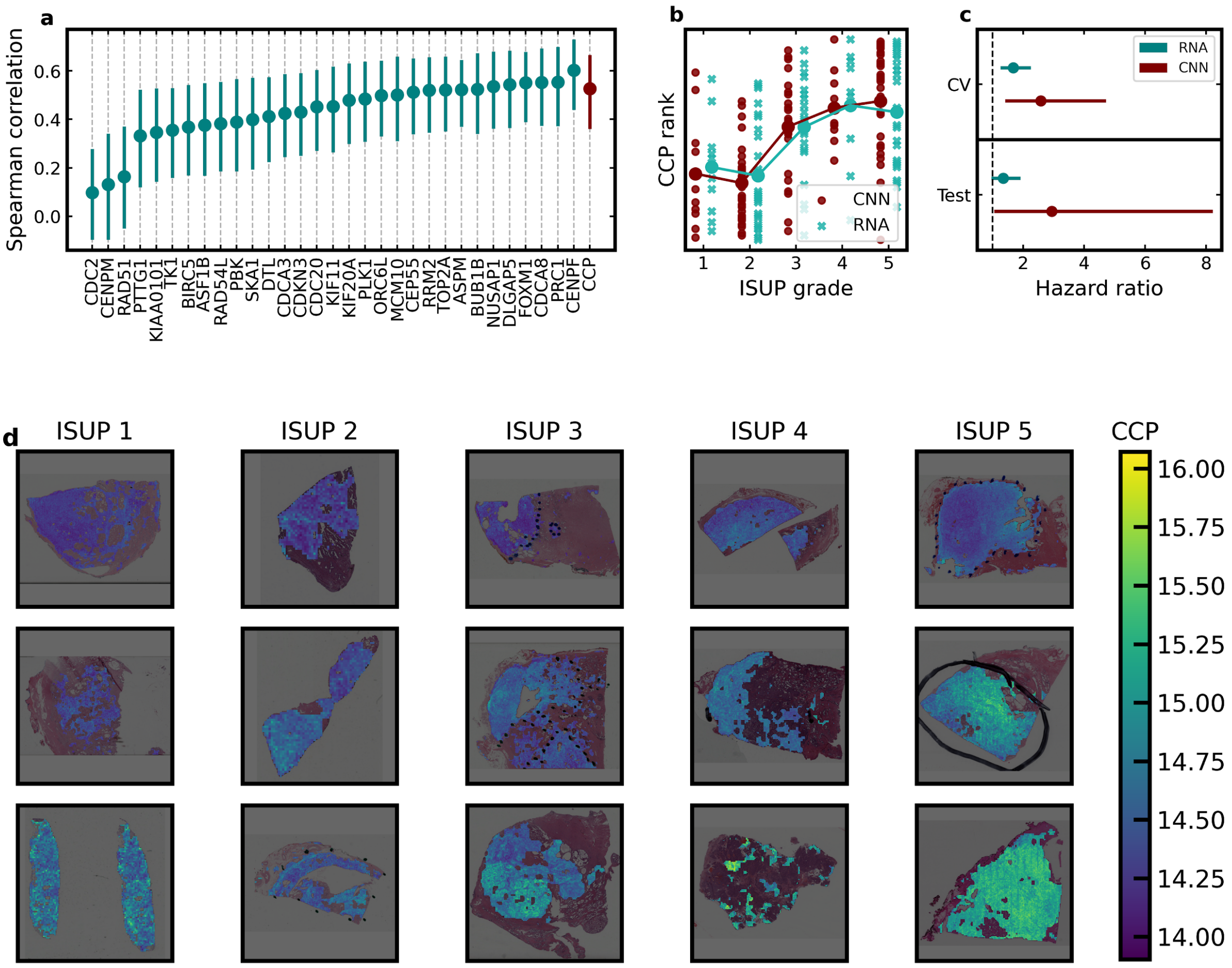

**Figure 5.** Comparison between the cell cycle proliferation (CCP) score based on RNA-seq and CNN predictions. **a)** Spearman correlation between sequenced and predicted gene expression in the test set with bootstrapped confidence intervals. **b)** Ranked CCP scores per ISUP grade both for RNA CCP as well as CNN CCP. **c)** Univariate hazard analysis for time to first biochemical recurrence for the RNA-seq based and predicted CCP score in the cross-validation data and the test set. The HR of the RNA-seq based CCP is 1.68 (1.256, 4.713) in the CV and 1.351 (0.956, 1.909) in the test data. For the predicted CCP, the respective HR values are 2.579 (1.412, 4.713) and 2.943 (1.055, 8.212) in the test set. **d)** Examples of WSIs per ISUP grade with overlaid local CCP score predictions.

# Supplementary methods and results

## Image preprocessing

The image preprocessing consists of tissue detection, tiling, normalisation and ROI identification through cancer detection. Tissue masks were generated by a logical AND operation between a mask based on Otsu thresholding[48] of the saturation channel of the WSIs, a mask that identifies regions with a hue of above 125° (on a scale up to 180°) and a local variance of Laplacian of above 20 to exclude blurry image regions. Morphological opening and closing was then used to remove salt-and-pepper noise from the masks. Visual inspection revealed that this led to unsatisfactory results for 9 of the TCGA WSIs. We therefore manually adjusted the filter criteria for these slides. The biopsy WSIs were then tiled at 10X magnification with 0.904 microns per pixel (mpp) into tiles spanning 598 x 598 pixels with a stride of 299 pixels along both dimensions. TCGA WSIs were tiled at 40X (0.252 mpp), 20X (0.504 mpp) and 10X (0.904 mpp) to a tile size of 500 x 500 pixels with a stride of 250 pixels along each dimension at 10X and 20X and 500 pixels at 40X. Tiles containing less than 50% tissue were excluded. In order to attenuate confounding variations in the WSIs that originate from different preparation protocols and different scanner properties, we colour normalised all tiles with the method described by Macenko *et al.*[49]. To this end, we computed an average stain vector from a random sample of tiles from the 5688 STHLM3 WSIs (compare next sections) in the development set of the biopsies. We then normalised all tiles from the biopsy development and test set and all TCGA tiles towards this reference stain vector.

## STHLM3 biopsy cohort for cancer detection

The dataset that we used to develop a cancer detection model consists of biopsies from the STHLM3 study, a prospective, population-based, screening-by-invitation study[30,31]. Each biopsy was annotated by a single pathologist, who delineated cancer regions and determined a Gleason grade according to the International Society of Urological Pathology (ISUP)[50]. Of the 10092 digitised biopsies, 1250 were excluded due to quality concerns. This resulted in 7211 biopsies from 1136 patients that were included in this study. 227 of these patients (19.98%, 1521 WSIs) were drawn randomly as a test set for the cancer detection model, stratifying for ISUP grade. The remaining 909 patients with 5688 slides were split into five cross-validation folds for hyperparameter tuning.

## Cancer detection model

Since our objective is to base the gene expression models only on areas of the WSIs that contain cancer tissue, we implemented a cancer detection model that is used as a pre-processing step. The cancer detection model was implemented as a tile-level classifier to generate binary cancer masks for the TCGA data set. Hyperparameters were optimised in a 5-fold cross-validation. This CNN model was based on the InceptionV3[51] architecture initialized with ImageNet[34] weights. The final fully connected layer of the Inception model was replaced by a drop-out layer with a 50% drop-out rate and a fully connected layer with two output neurons. Models were optimised using stochastic gradient descent (SGD) through 30 partial epochs, each consisting of 2500 batches with 24 samples (60,000 randomly sampled images per partial epoch with normal tissue and cancer tissue in equal proportion). During the hyperparameter optimisation, model performance was monitored during training on 500 validation batches that comprise 6,000 randomly sampled cancer tiles and 6,000 randomly sampled tiles containing normal tissue. As data augmentation, we deployed random rotations by multiples of 90°, random mirroring, random cropping to 500 x 500 pixels and color perturbations in the HSV color space with the same configuration as in Skrede et al.[52]. We found a learning rate of 1e-3 and multiplication of the learning rate with a factor of 0.2 after 10 and 20 partial epochs to be optimal. We then fitted 7 models to the entire development set, including the validation folds. Predictions on validation and test data are the average of the predictions of these 7 models per tile. Once predictions for all tiles were obtained, we removed salt-and-pepper noise from the resulting masks by applying morphological opening and closing.

## Cancer detection results

We first evaluated the performance of the cancer detection model in a biopsy test set consisting of 227 patients and 1521 WSIs. The tile level AUC is 0.87 as shown in Supplementary Figure 3a. We then obtained a slide level AUC by using the 99% quantile of all tile predictions per WSI as the free variable to compute an ROC, which is shown in Supplementary Figure 3b. This slide level AUC of 0.99 is comparable to the performance of previous publications about the STHLM3 cohort.

In order to evaluate the cancer detection performance in the TCGA cohort, we identified a subset of 77 WSIs from the TCGA training data that contain penmarks that annotate cancer regions and generated masks based on these penmarks. The tile level AUC as shown in Supplementary Figure 3c based on these WSIs is 0.832. Supplementary Figure 3d shows the calibration of the cancer detection model both for the STHLM3 test set and the annotated TCGA slides. Due to the difference in calibration between biopsies and resected prostates, we selected a threshold of 0.863 for cancer detection based on the TCGA tile level ROC with a TPR of 0.8 and an FPR of 0.321.

# Supplementary figures and tables

## Supplementary figures

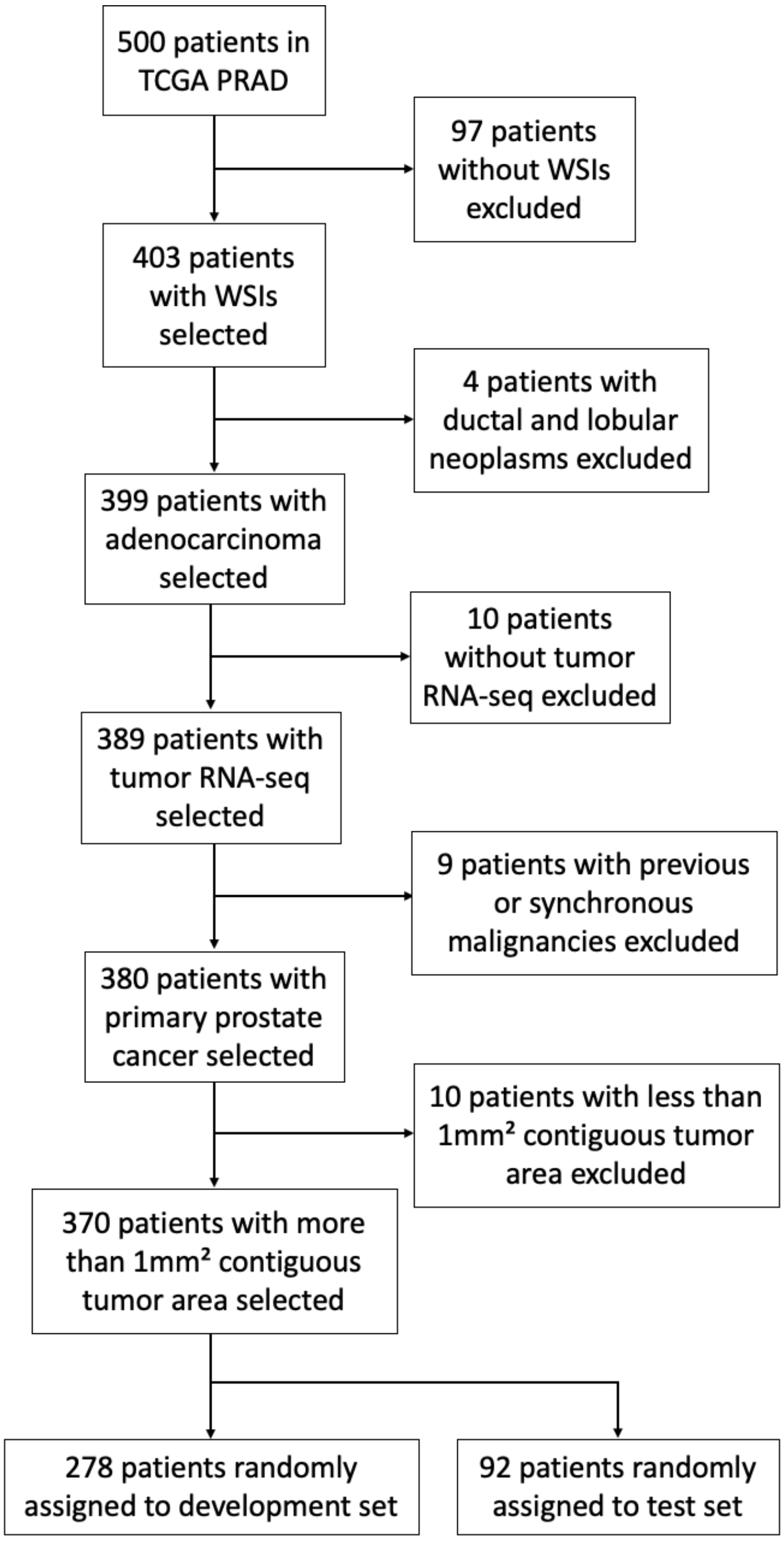

**Supplementary Figure 1.** CONSORT chart of the patient selection. Out of 500 patients in the TCGA PRAD cohort available through the GDC data portal, 403 patients with available WSIs were selected. Of those, 4 patients with ductal and lobular neoplasms were excluded, resulting in 399 patients with adenocarcinoma selected for this study. Of those,

389 had tumor RNA-seq available, which were subsequently selected. A further 9 patients were excluded due to previous or synchronous malignancies and associated systemic treatments. Of the remaining 380 patients, 10 with a contiguous tumor area of less than 1mm² were excluded, with 370 patients selected for inclusion. 92 of these were randomly assigned to the test set, stratified by ISUP and matching the age distribution, with 278 assigned to the development set.

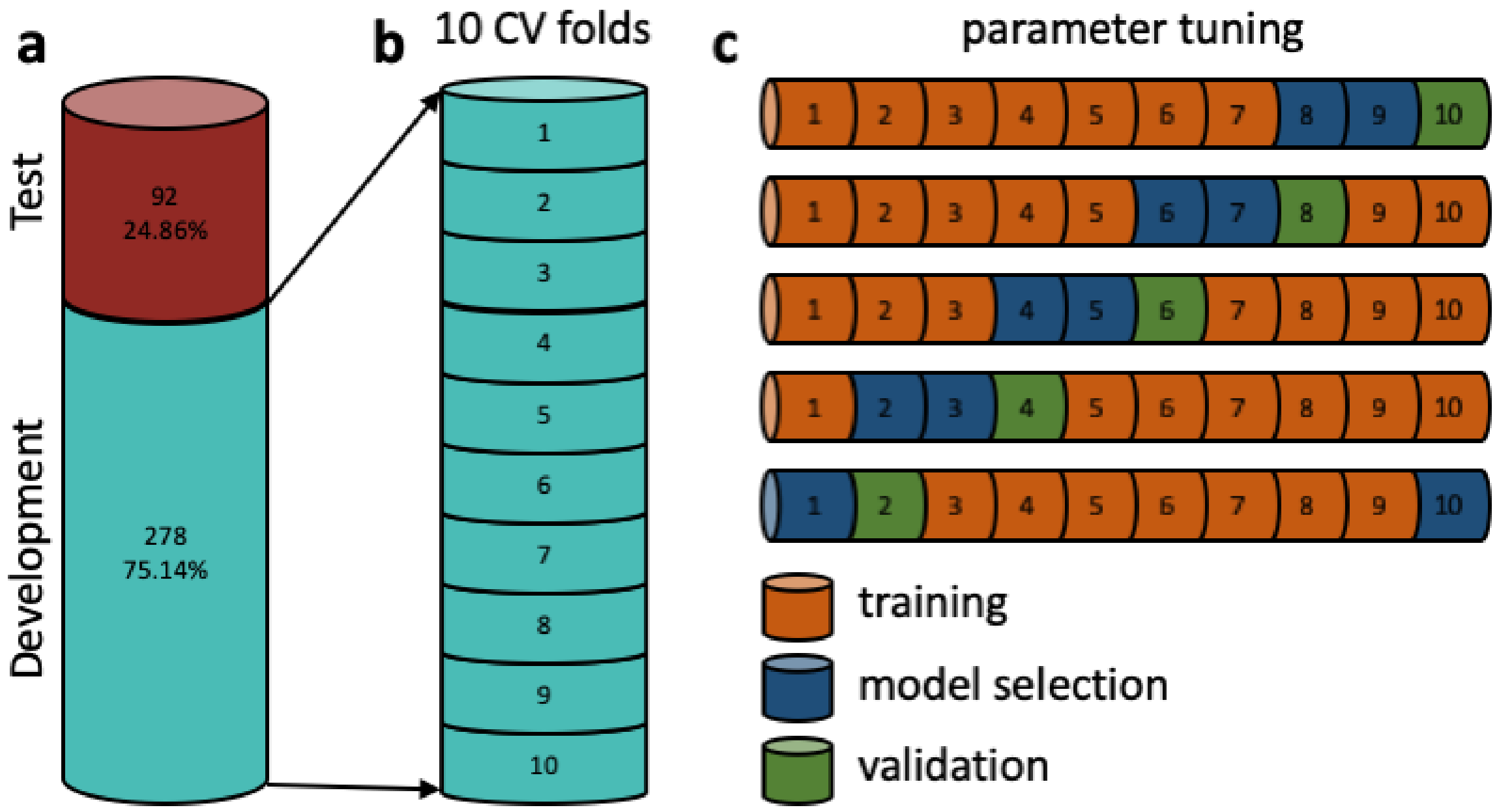

**Supplementary Figure 2.** Schematic of the validation procedure. **a)** The 370 included TCGA PRAD patients were split into a test set of 92 patients (24.86%) and a development set (278, 75.14%), stratified for ISUP grade and with matched age distributions. The development set was then further split into 10 folds, as in shown in **b)**. Hyperparameters were optimised in 5 fold combinations as depicted in **c)**, fitting models on 7 folds as training folds (orange) and using two model selection folds (blue) to estimate the performance of a hyperparameter set. Validation folds (green) were excluded during hyperparameter optimisation. The final models were then tuned in a 10-fold cross-validation with the folds in **b)**, including the model selection folds into the training data. Transcript predictions from the respective validation folds were then concatenated for the 278 patients of the development set in order to identify significantly associated transcripts (Spearman correlation, FDR adjusted p-value <0.0001). The significant associations were then validated in the held-out test set. To this end, predictions of the 10 models from the 10-fold cross-validation were averaged for each of the 92 cases in the test set.

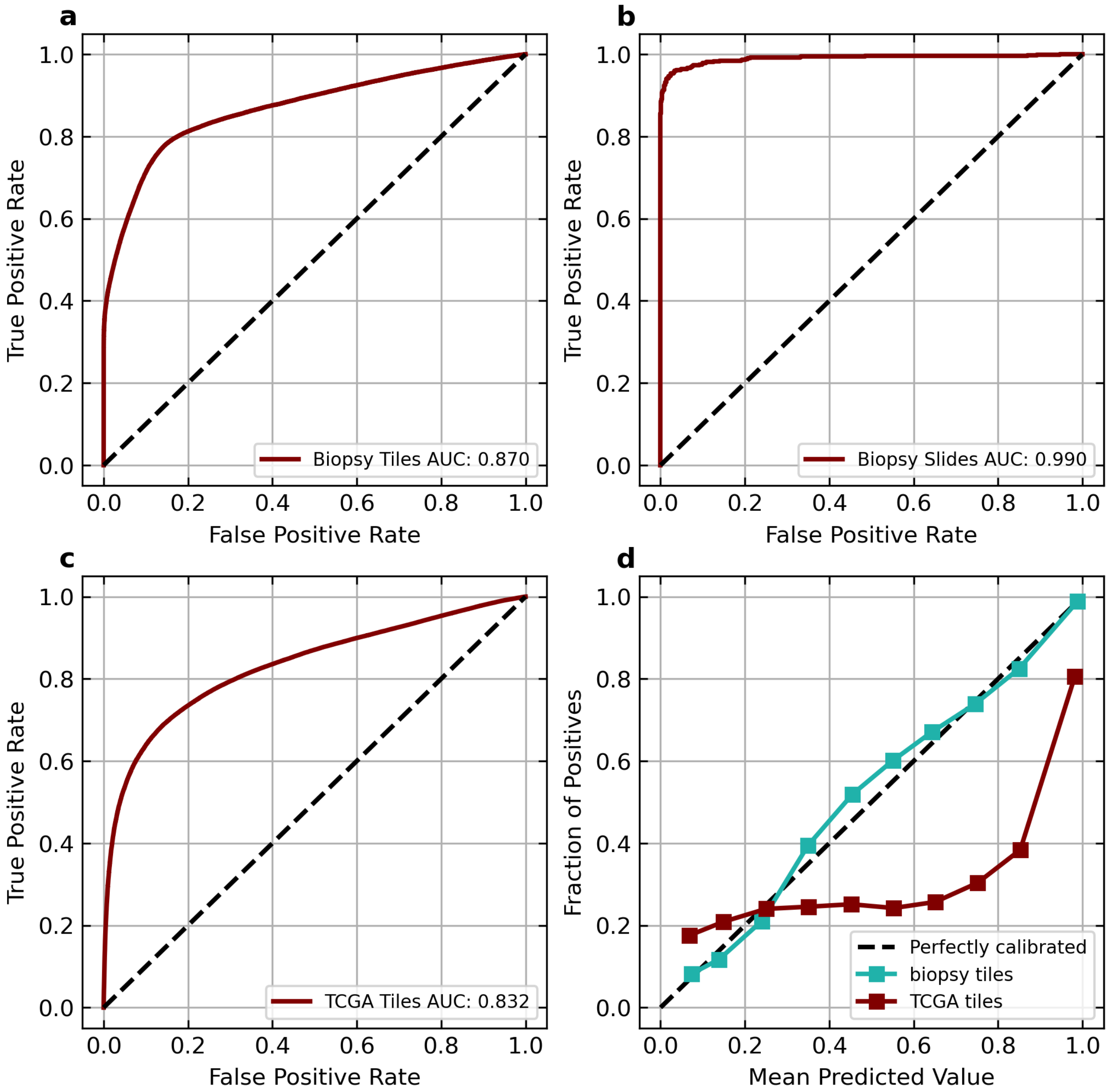

**Supplementary Figure 3.** Performance metrics of the cancer detection model. **a)** ROC for the tile-wise classification of benign against cancerous tissue in prostate needle biopsies with an AUC of 0.87 in the biopsy test set. **b)** ROC for slide-level classification of benign against cancerous biopsy cores where the tile-level predictions were aggregated to a slide-level prediction through their 99*th* percentile, with an AUC of 0.99 in the biopsy test set. **c)** ROC for the tile-wise classification of benign against cancerous tissue in resected sections for a subset of 77 TCGA slides from the development set that had penmark annotations for cancer regions available. The tile-level AUC is 0.832. **d)** Calibration of the cancer detection model in the biopsy test set and the TCGA development set.

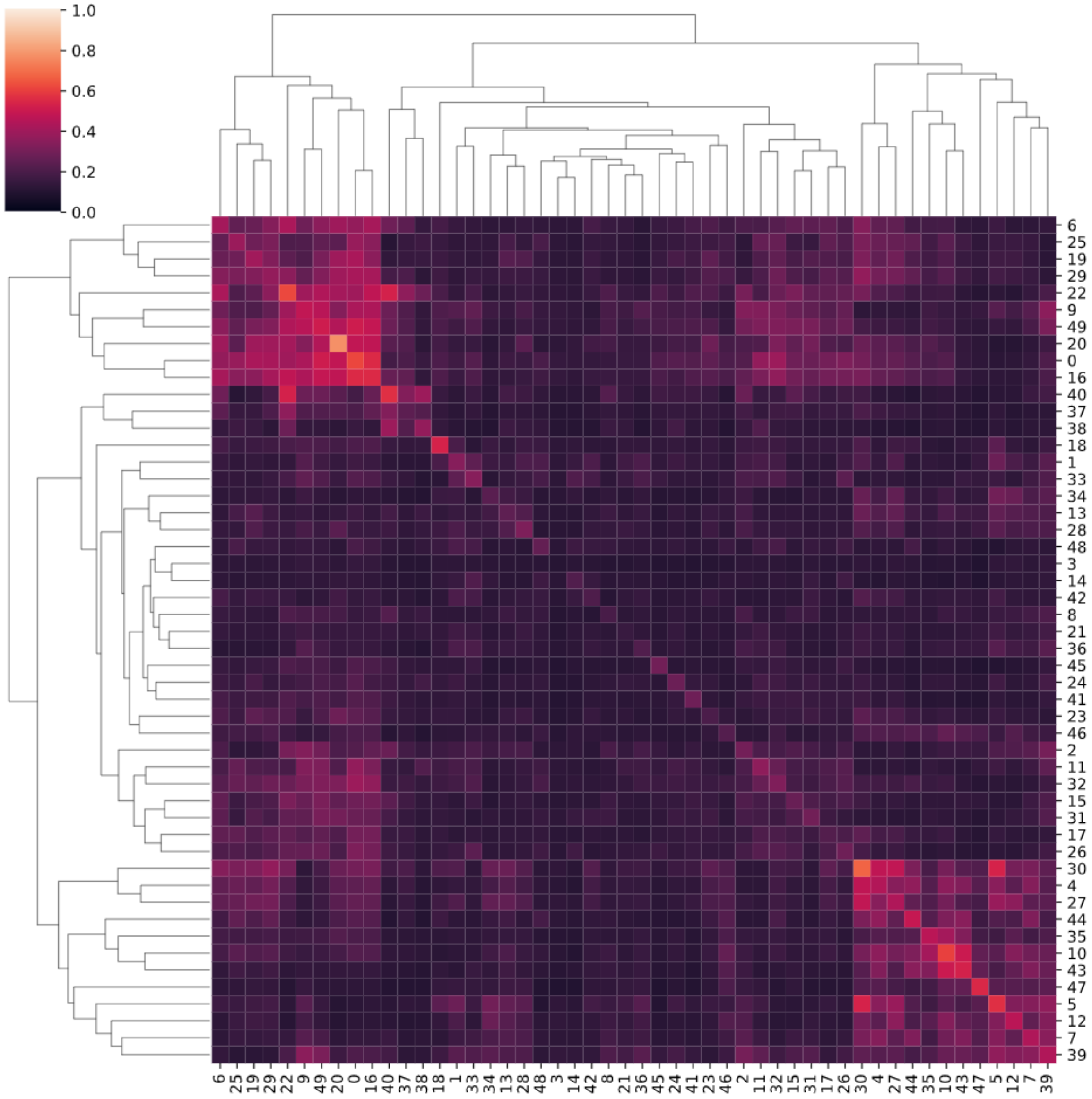

**Supplementary Figure 4.** Average absolute Spearman correlation for all gene pairs within a cluster or between two respective clusters.

# Supplementary tables

Supplementary Table 1. Correlations of CCP score.

| Gene | ensembl_id | Spearman correlation | Pearson correlation | Spearman ci_lower | Spearman ci_upper |
|---|---|---|---|---|---|
| RAD51 | ENSG00000051180 | 0.163 | 0.261 | -0.06 | 0.372 |
| PTTG1 | ENSG00000164611 | 0.332 | 0.445 | 0.12 | 0.518 |
| KIAA0101 | ENSG00000166803 | 0.347 | 0.448 | 0.144 | 0.527 |
| TK1 | ENSG00000167900 | 0.355 | 0.435 | 0.151 | 0.531 |
| BIRC5 | ENSG00000089685 | 0.368 | 0.483 | 0.16 | 0.546 |
| ASF1B | ENSG00000105011 | 0.376 | 0.461 | 0.172 | 0.547 |
| RAD54L | ENSG00000164080 | 0.383 | 0.402 | 0.187 | 0.545 |
| PBK | ENSG00000168078 | 0.389 | 0.471 | 0.191 | 0.568 |
| SKA1 | ENSG00000154839 | 0.399 | 0.469 | 0.197 | 0.576 |
| DTL | ENSG00000143476 | 0.412 | 0.507 | 0.222 | 0.579 |
| CDCA3 | ENSG00000111665 | 0.425 | 0.508 | 0.242 | 0.585 |
| CDKN3 | ENSG00000100526 | 0.431 | 0.545 | 0.244 | 0.588 |
| CDC20 | ENSG00000117399 | 0.453 | 0.514 | 0.267 | 0.609 |
| KIF11 | ENSG00000138160 | 0.453 | 0.403 | 0.259 | 0.621 |
| KIF20A | ENSG00000068796 | 0.479 | 0.487 | 0.294 | 0.632 |
| PLK1 | ENSG00000166851 | 0.484 | 0.552 | 0.303 | 0.635 |
| ORC6L | ENSG00000091651 | 0.499 | 0.595 | 0.329 | 0.644 |
| MCM10 | ENSG00000065328 | 0.501 | 0.516 | 0.31 | 0.663 |
| CEP55 | ENSG00000138180 | 0.513 | 0.568 | 0.33 | 0.658 |
| RRM2 | ENSG00000171848 | 0.52 | 0.534 | 0.352 | 0.657 |
| TOP2A | ENSG00000131747 | 0.521 | 0.441 | 0.35 | 0.66 |
| ASPM | ENSG00000066279 | 0.523 | 0.452 | 0.365 | 0.644 |
| BUB1B | ENSG00000156970 | 0.525 | 0.54 | 0.341 | 0.674 |
| NUSAP1 | ENSG00000137804 | 0.536 | 0.533 | 0.362 | 0.678 |
| DLGAP5 | ENSG00000126787 | 0.544 | 0.557 | 0.368 | 0.686 |
| FOXM1 | ENSG00000111206 | 0.552 | 0.597 | 0.395 | 0.68 |
| CDCA8 | ENSG00000134690 | 0.552 | 0.565 | 0.374 | 0.698 |
| PRC1 | ENSG00000198901 | 0.554 | 0.575 | 0.377 | 0.701 |
| CENPF | ENSG00000117724 | 0.602 | 0.608 | 0.439 | 0.73 |
| CCP | CCP | 0.527 | 0.558 | 0.357 | 0.665 |

## Supplementary Table 2. Hyperparameters CNN

| **Hyperparameter** | **Considered** | **Selected** |
|---|---|---|
| Model architecture | InceptionV3, ResNet18 | ResNet18 |
| Learning rate | 1e-3, 1e-4, 1e-5 | 1,00E-03 |
| Learning rate decay | 0.95 | 0.95 |
| No. of train epochs | up to 50 | 25 |
| images/epoch | 24000 | 24000 |
| batch size | 144 | 144 |
| image size (after crop) | 440 x 440 | 440 x 440 |

Supplementary Table 3. Hyperparameters LGBM

| Hyperparameter | Considered | Selected |
|---|---|---|
| learning_rate | 0.01, 0.05, 0.1 | 0.1 |
| max_depth | 12, 24 | 12 |
| num_leaves | 12, 24 | 24 |
| feature_fraction | 0.05, 0.1, 0.2 | 0.2 |
| bagging_fraction | 0.2, 0.4, 0.6 | 0.2 |
| bagging_freq | 1 | 1 |
| num_iterations | up to 500 | 80 |
| early_stopping_rounds | 15 | n.a. |